\documentclass[letterpaper, 10 pt, conference]{ieeeconf}  

\IEEEoverridecommandlockouts                              

\usepackage[bookmarks=true]{hyperref}

\usepackage{array,multirow,graphicx}
\usepackage{amsmath,amssymb}
\usepackage{wrapfig}
\usepackage{color}
\usepackage{ulem}

\usepackage{subcaption}
\usepackage{algorithmicx}
\usepackage{algorithm}
\usepackage{algpseudocode}
\usepackage{booktabs}
\usepackage[bookmarks=true]{hyperref}
\usepackage{wrapfig,lipsum,booktabs}

\usepackage{siunitx}

\title{\LARGE \bf
CLAMGen: Closed-Loop Arm Motion Generation \\ via Multi-view Vision-Based RL
}

\author{Iretiayo Akinola\thanks{$^*$Equal Contribution}$^{*1}$, Zizhao Wang$^{*1}$, and Peter Allen$^{1}$
\thanks{This work was supported in part by a Google Research grant and National Science Foundation grant IIS-1527747.
} 
\thanks{$^{1}$Department of Computer Science, Columbia University, New York}%
}

\begin{document}

\maketitle
\thispagestyle{empty}
\pagestyle{empty}

\begin{abstract}

    We propose a vision-based reinforcement learning (RL) approach for closed-loop trajectory generation in an arm reaching problem. Arm trajectory generation is a fundamental robotics problem which entails finding collision-free paths to move the robot's body (e.g. arm) in order to satisfy a goal (e.g. place end-effector at a point).
    While classical methods typically require the model of the environment to solve a planning, search or optimization problem, learning-based approaches hold the promise of directly mapping from observations to robot actions.
    However, learning a collision-avoidance policy using RL remains a challenge for various reasons, including, but not limited to, partial observability, poor exploration, low sample efficiency, and learning instabilities.
    To address these challenges, we present a residual-RL method that leverages a greedy goal-reaching RL policy as the base to improve exploration, and the base policy is augmented with residual state-action values and residual actions learned from images to avoid obstacles. Further more, we introduce novel learning objectives and techniques to improve 3D understanding from multiple image views and sample efficiency of our algorithm.
    Compared to RL baselines, our method achieves superior performance in terms of success rate. 
\end{abstract}


\section{Introduction}
Reinforcement learning is increasingly being applied to robotic applications such as aerial navigation, locomotion, and manipulation. A large component of robotic manipulation is trajectory planning which involves finding a collision-free path from a start configuration to a goal point. Classical approaches such as sampling-based methods \cite{lavalle1998rapidly,karaman2011sampling,kavraki1998probabilistic} and optimization-based ones CHOMP~\cite{ratliff2009chomp}, STOMP~\cite{kalakrishnan2011stomp} are able to solve the problem given enough time; they tend to trade computational speed for optimality \cite{bekris2019asymptotically}.
These classical methods require a model of the environment to solve the motion planning problem, continuous changes in the dynamic environments require constantly updating the environment model used for planning. On the other hand, learning-based approaches present a way to generate trajectories directly from sensory inputs.
More recently, learning-based methods have become attractive for motion planning because of their constant computational cost which is simply querying a pre-trained network and does not scale with environment complexity.

\begin{figure}[t]
\centering
	\includegraphics[width=1\linewidth,keepaspectratio]{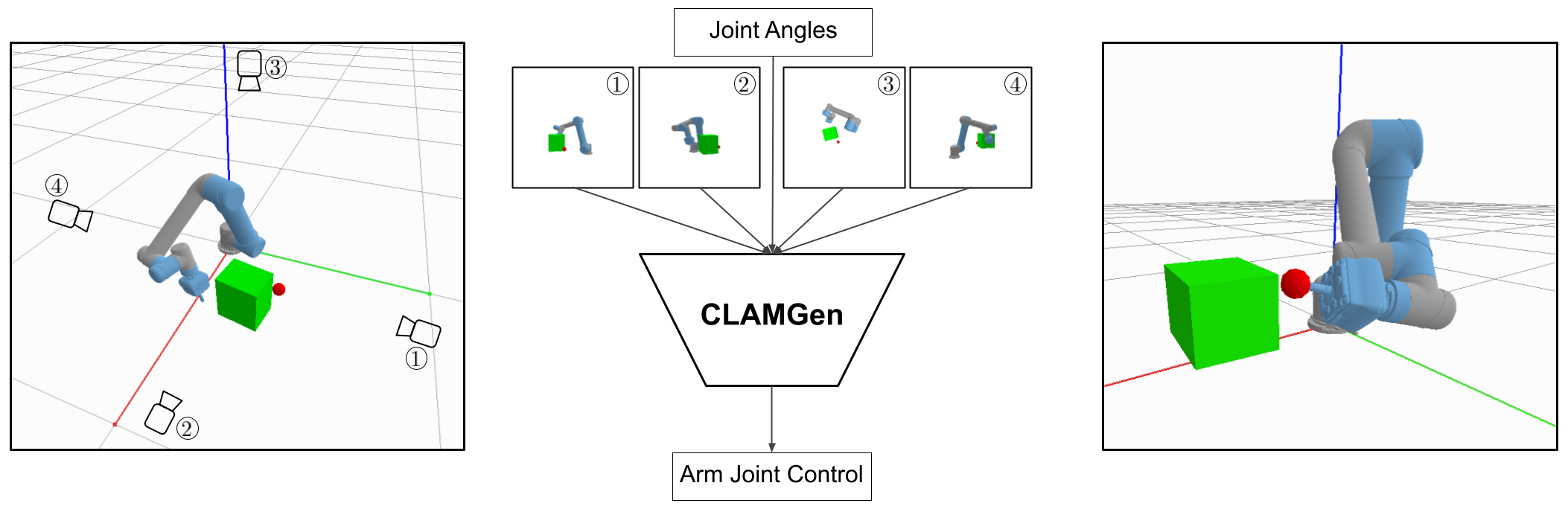}
\caption{ Vision-Based Closed-Loop Trajectory Generation for a 6-DOF Robot Arm: How can the robot efficiently move its arm to place its end effector at target locations (red) without hitting obstacles (green)? It needs to learn 3D workspace awareness from multiple-views to generate valid motions. Using a residual RL formulation, CLAMGen learns a reactive policy that uses multi-view image observations with the robot joint angles to generate arm motion that moves the end-effector to the goal location. Videos showing generated arm trajectories can be found at our project website \url{http://crlab.cs.columbia.edu/CLAMGen}.}
\label{fig:environments}
\end{figure}

Learning a collision-avoidance behavior presents a number of unique challenges. First, a rich observation is required to capture the robot and scene geometries in a way that reveals the robot's contact with the environment to prevent collision. The common practice of using a static single-view camera would be insufficient in this case. Second, adding more camera views increases the size of the observation space and consequently exacerbate the sample-complexity issues. Third, it is difficult to derive an adequate reward function that achieves the obstacle avoidance behavior with issue. The sparse reward functions that penalize collision and reward reaching the goal have poor exploration properties. For example, the collision penalty can make the agent super conservative, explore less and unable to reach the goal. Without the penalty, the agent exploits collision with the environment to reach the goal and consequently is unable to achieve collision avoidance. This work tackles these challenges of applying RL  to obstacle avoidance by using residual learning.

Path planning (arm motion generation in our case) can be thought of as the motion of a robot through a potential field where there is an attractive potential at the goal and a repulsive potential around obstacles \cite{khatib1986real}. While the attractive potential can be captured by the distance from the robot to the goal, the obstacle repulsive potential is a more complicated function of the obstacle configuration and geometry as well as of the robot configuration and robot geometry. Obtaining these two potentials and combining them appropriately can be challenging.
In this work, CLAMGen learns both potential fields in a two-stage process. First, we obtain the attractive potential in the obstacle-free low-dimensional robot configuration space using a goal-based reward function to capture the attractive potential in the form of a Q-value function. Second, we learn a vision based residual to modify the obstacle-free Q-function obtained from the previous stage. This vision-based residual Q-value is a function of images from multiple camera views and the policy obtained from the combined Q-function reasons about the obstacle avoidance while trying to reach the target. The output of our residual policy is added to the output of the base policy. This modulates the base action to obtain a collision-avoidance behavior.

Learning in simulation is useful and valuable for obstacle avoidance since collision can be costly in real world.  We leverage this in two ways. First, we allow collision in the early parts of the training to encourage full exploration and improved learning of the dynamics of the environment. As learning proceeds, we increasingly penalize collision to obtain the desired collision avoidance behavior. Secondly, we can control the collision threshold in simulation which when increased would obtain a safer policy. In other words, penalizing the robot for being close the obstacle can make it less likely to touch it after training. Our work focuses on leveraging simulation to obtain an obstacle avoidance behavior, existing sim-to-real techniques can be used to demonstrate our policy on a real robot. We leave sim-to-real adaptation of our method as part of future work.


In summary, our key contributions in this work include:
\begin{itemize}
    \item an adaptive, closed-loop vision-based policy for obstacle avoidance capable of reaching static and dynamic targets.
    \item a novel residual Q-learning formulation that enables sample-efficient reinforcement learning of the obstacle avoidance arm movements; our method does not require any expert demonstrations.
    \item we show that it is difficult to generate collision-free arm trajectories using only joint information and single-image views.
    \item an improved 3D awareness of the robot agent using across-view constrastive loss and collision dynamics loss.
    \item we leverage simulation to effectively apply reversible-HER (forward and inverse hindsight experience replay \cite{andrychowicz2017hindsight}) to learning from images.
    
\end{itemize}


\section{Related Work}

\subsection{Learning-Based Planning}
Recent works \cite{tamar2016value}\cite{srinivas2018universal} \cite{qureshi2018motion} \cite{ichter2018learning} \cite{ichter2018robot} \cite{choudhury2017data} \cite{kuo2018deep} have explored learning-based approaches to solving motion planning problems. While some employ a supervised imitation learning approach to train a model to mimic example paths, others use reinforcement learning techniques to learn path generation by trial and error. Some focus on grid-based planning for discrete worlds \cite{le2018hierarchical} \cite{panov2018grid} while others do continuous space planning \cite{tamar2016value, jurgenson2019harnessing}. Another class of learning-based methods do planning in a learned latent space \cite{ichter2018robot}\cite{hafner2019learning}. These methods propose converting the high-dimensional representation of obstacles such as images into a low-dimensional latent space that captures connectable collision-free manifold of the environment dynamics. Planning is done by sampling this latent space conditioned on the start and goal locations.

A big challenge with applying learning-based methods to path-planning is non-generalizability; \cite{dhiman2018critical} highlighted the difficulty of learned methods to generalize to the full path planning problem where the environment obstacles, start and goal locations are completely randomized.
Because of this limitation, hybrid methods are developed to combine learning-based methods with conventional sampling based methods \cite{qureshi2018motion}\cite{ichter2018robot}.
Nevertheless, the short inference time of neural networks motivates improvement on learning-based motion planning modules to further close the generalizability gap.
For example, very recent work \cite{ha2020learning} demonstrated the value of using learning for multi-arm planning using propriceptive information. Their multi-agent RL approach achieved superior closed-loop performance leveraging fast inference time of neural networks. This further motivates our approach to leverage learning in obtaining a vision-based policy.

In contrast to existing works, we introduce a fully-learned cross-modal residual policy learning where a vision-based residual policy is learned on top of low-dimensional state-space. Our approach obtains a natural decomposition of the motion planning problem into target reaching and obstacle avoidance. The target reaching problem can be solved using low-dimensional observation while the obstacle avoidance requires visual information. After learning a low-dimension reaching policy, we use residual learning to obtain a final policy that avoids obstacles while reaching the target.

\subsection{Vision-Based Reinforcement Learning}

Recent works have shown that a useful dynamics model can be learned from visual inputs  using reinforcement learning~\cite{hafner2018learning}\cite{srinivas2020curl}. 
They are able to overcome the dimensionality challenge of images using auxilliary image reconstruction objectives, latent dynamics loss to encourage consistency in the latent space, and contrastive loss to also impose structure on the latent space among others.
We observe that most of these policies were learned on a single image observation which is insufficient to obtain a good 3D obstacle-avoidance behavior.
Our work focuses on the use of multiple camera-views to enable learning an obstacle avoidance task that require 3D understanding.

Building on existing works, we extend CURL to multi-view settings by using across-view contrastive loss to make latent representation similar across views. In addition, we do auxiliary image prediction of a hidden view from a subset of the camera views. These two enhancements encourage the network to form an implicit association among the views to aid 3D understanding of the environment. In addition, the auxiliary image reconstruction is useful for the unsupervised adaptation of our policy to a new environment.
A previous work \cite{byravan2018se3} learned a visuomotor reaching policy for controlling a full-bodied robot manipulator in joint space to reach a goal point in task-space in an obstacle-free environment; this was accomplished using a single depth image. We show that a single-view is insufficient for 3D obstacle avoidance and demonstrate how to effectively overcome this limitation with more camera-views and in a sample-efficient manner. 
Our work focuses on learning closed-loop motion generation policy from multiple image-views to reach target locations while avoiding obstacles in the scene.


\subsection{Sample Complexity Challenge}
End-to-End reinforcement learning from images is particularly challenging in terms of sample complexity. To overcome this, some works use expert policy to bootstrap the learning process \cite{kalashnikov2018qt,nair2018overcoming,akinola2020learning}, some use hind-sight experience replay \cite{andrychowicz2017hindsight}, some use auxiliary tasks \cite{jaderberg2016reinforcement,yarats2019improving}. Our work combines a variant of residual learning and hindsight replay.

\subsubsection{Residual Policy Learning}
Recent \textit{Residual Policy Learning} works \cite{silver2018residual, johannink2019residual} present an effective approach to improve on a baseline nondifferentiable policies using model-free deep RL in continuous action spaces. They showed that learning a residual improves sample complexity of deep RL and demonstrated the effectiveness of the approach to a number of low-dimensional non-visual tasks. \cite{zeng2020tossingbot} learned a vision-based residual policy on top of a physics-based controller.
In contrast to these existing works, our method learns a residual policy for a target task (image-based obstacle avoidance) using base policy for a different but related task (joint-based arm control). We obtain a vision-based residual policy from multi-view camera views, a very high-dimensional state-space to augment a learned low-dimensional joint-space policy. In addition, our approach also enables jointly updating both base and residual policies.

\subsubsection{Image-Based Hindsight Experience Replay}
Hindsight Experience Replay (HER) \cite{andrychowicz2017hindsight} has been shown to improve sample efficiency of goal-based off-policy reinforcement learning algorithms without a need for expert demonstration. HER achieves this by re-labeling the final state of failed attempts as the originally intended goals. This technique has been successfully applied to low-dimensional learning problems however, this direct re-labelling approach does not directly apply to image-based tasks where the visual observations would look differently for a reimagined goal \cite{sahni2019addressing}. Since the goals must be inferred from the image, image observations of relabelled trajectory have to be regenerated or re-rendered. \cite{sahni2019addressing} addressed this by training a GAN to regenerate the new image observations given a reimagined goal. This was effective especially for a turtlebot navigation task where the goal is almost always in site.
In our case, image observation is susceptible to occlusion from different views at different time-steps during the trajectory which would be difficult for GAN hallucination to recreate with high fidelity. To address this, we leverage simulation to regenerate visual observation for relabelled trajectories. This is done on a separate process allowing learning to proceed without delay. Furthermore, we relabel the reversed version of the collision-free subtrajectories as additional high-quality examples.


\section{Preliminaries}

\textbf{RL Formulation.}
We formulate arm trajectory generation as a multi-goal RL problem with the objective of learning a \textit{universal policy}\cite{schaul2015universal} to reach multiple goals.
The Markov decision process (MDP) comprises: the observation space $S$, an action space $A$ and a transition function $T : S \times A \rightarrow \Delta S$. Let the space of goals be $G$, the goal-dependent reward function is given as $R : S \times A \times G \rightarrow \Delta \mathbb{R}$.
A policy $\pi : S \times G \rightarrow A$ is learned to maximize the cumulative reward $\sum_{t=0}^{\infty} \gamma^t R(s_t, a_t, g)$, where $0 \le \gamma \le 1$ is the discount factor.
The state-action value function of the policy $\pi$ is given as: $Q^{\pi}: S \times A \times G \rightarrow \Delta \mathbb{R}$ which capture the expected discounted cumulative reward from a given state and goal while acting according to policy $\pi$.
The observation space $S$ consists of the current joint angles and image-views with cameras overlooking the environment. The continuous action space $A$ correspond to joint velocities for each joint.
A policy $\pi_{\theta}$ is learned to select actions in the MDP.

\textbf{Robot Simulation Setup.}
We use the Universal Robot UR5 arm\footnote{ \url{https://www.universal-robots.com/products/ur5-robot/}} fitted with an OnRobot VGC suction gripper\footnote{ \url{https://onrobot.com/en/products/vgc10}} as the robot platform in our experiments. We set this robot up in the  PyBullet \cite{coumans2016pybullet} simulator and place four cameras in the environment overlooking the robot workspace (Figure~\ref{fig:environments}). We randomly place a number of obstacles (green) in the workspace; the obstacles can be cube or cylinder with varying sizes.

\begin{figure*}
\centering
	\includegraphics[width=1\linewidth,keepaspectratio]{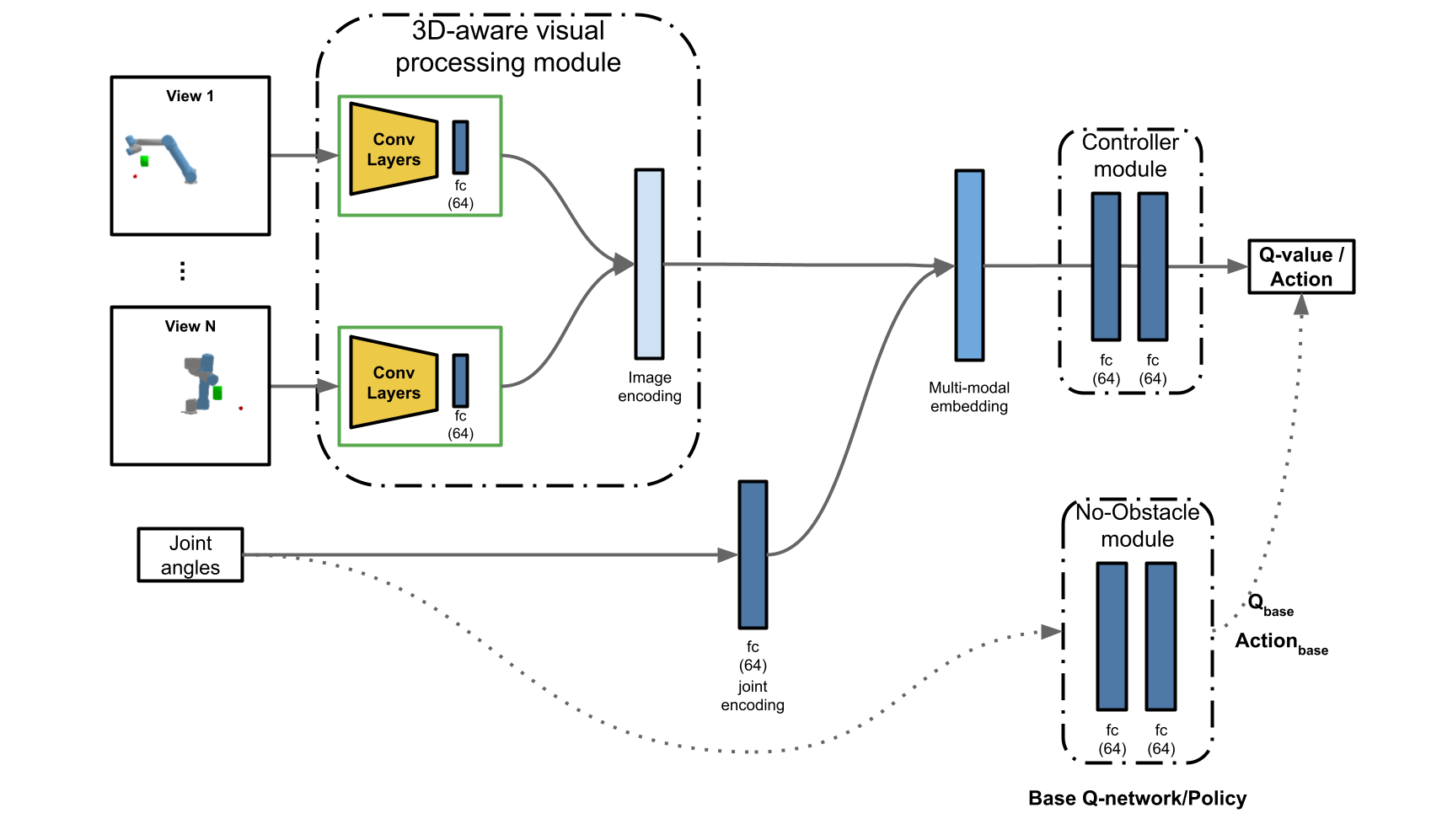}
\caption{CLAMGen Architecture: The visual processing module extracts 3D-aware image features which are merged with the features from low-dimensional input to form a multi-modal embedding. The fused embedding goes into the controller module of a few fully-connected (FC) layers to produce the action output that is added to the output of the base policy. The same architecture is used for the Q-value network where the output is combined with output of the base Q network (joint-based no-obstacle module). The dotted arrow indicates the path through the frozen base model whose parameters are not updated during training.}
\label{fig:network_architecture}
\end{figure*}

\section{Arm Motion Generation}
CLAMGen uses RL to obtain the arm motion generation by learning two potential fields: a field that attracts the robot end-effector to the goal location and a field that repels the robot from obstacles. In the RL context, CLAMGen models each field as a Q function that captures the value of each state-action pair given a goal and the obstacle configuration. The first Q function captures the value of state and action pairs if there are no obstacles in the environment. The second Q function is the residual that captures the change in the value of state-action pairs that give the obstacle avoidance field. The policy, which modulates the base action, is trained to maximise the combined Q-function. To do this, our algorithm builds on a state-of-the-art off-policy RL algorithm-- soft actor critic (SAC) whose objectives are given as:
\begin{align}
    \textnormal{critic: }  J_Q &= \mathop{\mathbb{E}} \left[ \frac{1}{2}(Q_\theta(s_t, a_t) - (r_t + \gamma V_{\bar{\theta}}(s_{t+1}))^2 \right] \label{eq:critic_loss}\\
     V_{\bar{\theta}}(s_{t+1}) &= \mathop{\mathbb{E}}_{a' \sim \pi_\theta} \left[Q_{\bar{\theta}}(s_{t+1}, a') - \alpha \log\pi_\theta(a'|s_{t+1})\right]\nonumber\\
    \textnormal{actor: }
     J_\pi &= \mathop{\mathbb{E}}_{a \sim \pi_\theta}\left[Q_\theta(s, a) - \alpha \log\pi_\theta(a|s)\right],
\end{align}
where $\theta$ are the parameters to optimize and $\bar{\theta}$ are delayed parameters.

\subsection{Obstacle-Avoidance via a Learned Residual}
For arm-motion generation, CLAMGen decomposes the Q-function $Q_{\theta}$ into two i.e. $Q_{\theta} = Q_{\theta_1} + Q_{\theta_2}$, where $Q_{\theta_1}$ capture the motion of the arm when no obstacle is presented while $Q_{\theta_2}$ handles the residual when obstacles are present.
\begin{align}
    \label{residual_equation}
    Q_\theta(s_t, a_t, g_t) = Q_{\theta_1}(s_{1_t}, a_t, g_t) +Q_{\theta_2}(s_{2_t}, a_t, g_t)
\end{align}
Note that $g_t$ dependence on $t$ captures the case of a dynamic target and the state space for each of the two $Q$s is a subset of the full observation space i.e. $s_{1_{t}}, s_{2_{t}} \subseteq s_{t}$ and $s_{1_{t}} \cup s_{2_{t}} = s_{t}$. For example, without obstacles, the state space $s_{1_{t}}$ of joint angles and goal location is sufficient. For obstacle avoidance, the state space $s_{2_{t}}$ also includes images.

The two Q functions and their accompanying policies are obtained in a two stage process. In the first learning stage, a policy $\pi_{\theta_1}$ is learned to move the arm in joint space to a Cartesian target position. This uses low-dimensional components of the observation space to learn a joint-based policy and Q-function that captures the inverse kinematics of the arm.
A simple reward function coupled with our reversible HER (Section \ref{reversible_her}) is able to learn a goal reaching behavior. Specifically, our reward only has a constant penalty for every episode step until the goal is reached. 
This reward function is maximised if the agent is able to reach the goal as fast as possible.
At the end of the first stage, we obtain an arm reaching policy for an environment without obstacle and an accompanying Q-value function.

To learn the obstacle avoidance behavior, the second stage utilizes the full observation space (including multi-view images with the joint angles) to obtain a residual that augments the joint-only policy and value function from the previous stage. The simple reward function of the first stage is retained enabling the robot collide freely especially in the early stage of learning. This improves exploration, enables the robot learn about the dynamics of the environment while still allowing the agent to reach the goal. At this point, the agent is able to reach the goal while rubbing against obstacles. As learning proceeds, the collision penalty gradually ramps up so that the agent adjusts it's behavior of reaching the goal without touching the obstacles. 
From the collision penalty, $Q_{\theta_2}$ captures a repulsive field for obstacles, and the new policy $\pi_{\theta_2}$ learned from the combined $Q_\theta$ would reach the goal while avoiding obstacles. As the joint-based policy serves as a good starting point for the new policy to adapt, there is a warm-up process in the beginning of the second stage. During warm-up, the new policy is optimized by maximizing $J_{\pi_{\theta_2}} = \mathop{\mathbb{E}}_{a \sim \pi_{\theta_2}}\left[Q_{\theta_1}(s, a) - \alpha \log\pi_{\theta_2}(a|s)\right]$ to clone $\pi_{\theta_1}$. Meanwhile, $Q_{\theta_2}$ is updated using the same critic loss as Eq \ref{eq:critic_loss}.

\subsection{Learned 3D fusion for Obstacle-Avoidance}

To enable efficient learning from multiple views, CLAMGen introduces the following techniques to improve 3D understanding and enable obstacle avoidance behavior.

\begin{itemize}
    \item \textbf{Contrastive loss}: CLAMGen uses the contrastive loss from CURL \cite{srinivas2020curl} to learn an encoding that can discriminate instances in the image.
    In detail, the query and positive target is obtained from two different augmentations (random crops) of the same image, and negative targets are augmentations gotten from other images. The loss trains the encoder to generate similar embeddings for the query and positive target, and distinct ones for negative targets.
    In addition, this pairwise loss is applied to reduce the distance in latent space between different views on the same scene and increase the distance on different scenes; this encourages the network to make associations across multiple viewpoints.
    \item \textbf{Collision-Dynamics loss}: given the current latent $z_t$ and an action $a_t$, CLAMGen predicts if a collision occurs after executing the action. This loss encourages the latent space to encode the dynamics of the environment.
    \item \textbf{View-Dropout}: We extend the idea from \cite{akinola2020learning} to randomly drop one or more of the views from cameras with a certain probability. This forces the agent to use every view and avoid over-dependence on a certain view.
\end{itemize}

Figure~\ref{fig:network_architecture} shows the CLAMGen architecture used for the Q-value function and policy. The 3D-aware visual processing module consists of 4 convolutional layers for each view, the output of which is merged and passed through two dense layers to produce visual encoding. Low dimensional inputs like joint angles are passed through a dense layer to get a joint encoding. The controller module which has two dense layers processes the merged multi-modal encodings into the policy output or Q-value. For each layer, we used the ReLU activation function to provide nonlinearity. For each stage, the Q-function was trained using the clipped double-Q trick \cite{van2016deep}. Additionally, we have auxiliary losses that capture the contrastive loss and the collision dynamics loss described above.



\subsection{Reversed Vision-Based Hindsight Experience Replay}
\label{reversible_her}
During reinforcement learning of an obstacle avoidance skill, the robot is bound to collide with obstacle many times, especially during the early stages of the trial-and-error based approach. This can have significant effect on the sample efficiency of the learning process since there would be few positive examples without collision. To improve the sample complexity, CLAMGen uses Hindsight Experience Replay (HER) to relabel the failure cases as success by ascribing the state before collision as the originally intended goal of the trajectory rollout. 
Importantly, our algorithm also use the reversed version of the relabelled collision-free trajectory by reversing the state-action sequence starting from the state just before collision. This provide examples of how to move away from obstacles in addition to further improving sample efficiency.
Given that the goal is inferred from the images, the images of the relabelled trajectory are re-rendered in simulation.



\section{Experimental Results}
\label{sec:result}


This arm trajectory generation task has two main components: 1) learn the forward/inverse kinematics from data to move to the goal. 2) learn to recognize and avoid obstacles. Our experiments are directed toward testing the performance of our method on these two components.

\subsection{Arm Reaching (No Obstacle)}
\label{sec:No_Obstacle}

\begin{figure}[h]
\centering
	\frame{\includegraphics[width=0.45\linewidth,keepaspectratio]{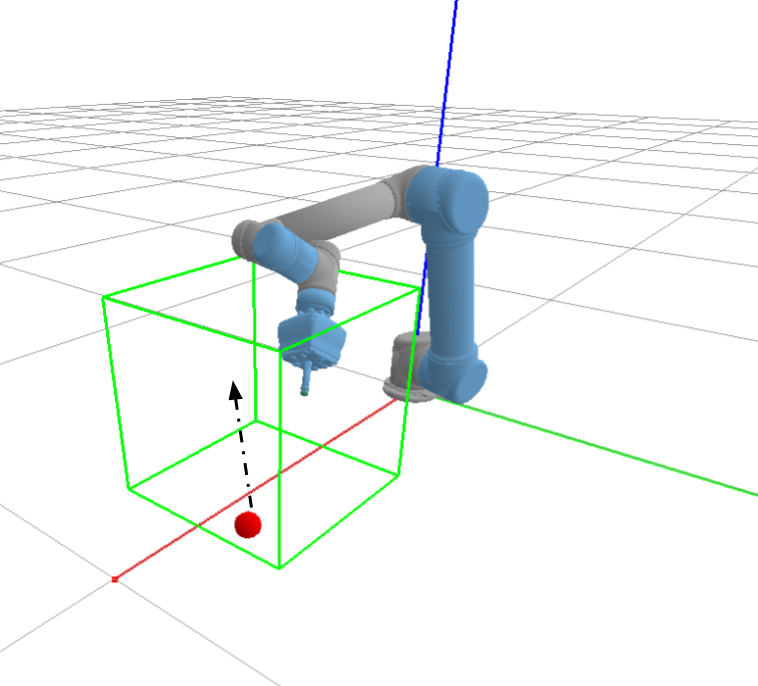}}
	\frame{\includegraphics[width=0.489\linewidth,keepaspectratio]{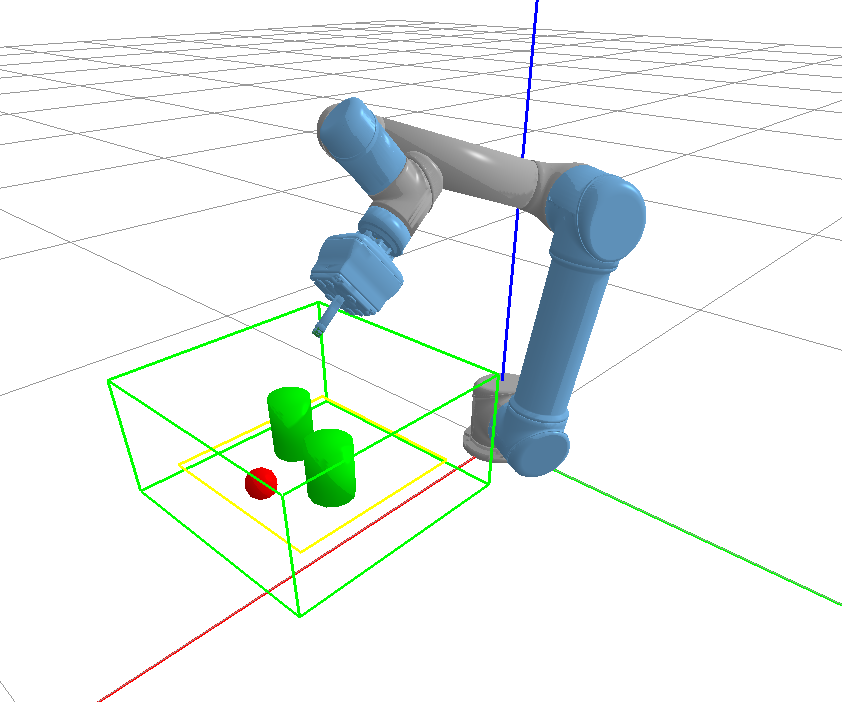}}
\caption{\textbf{Left}: Dynamic Reaching Task. \textbf{Right}: Obstacle Avoidance Task. Green box indicates the workspace from which goal is sampled, yellow box indicate the region from which obstacles are randomly placed.}
\label{fig:task_workspace}
\end{figure}

The first task is an arm-reaching task without obstacles. Shown in Figure \ref{fig:task_workspace} (left), the task workspace is a box with each side length of $0.4$m centred at ($0.6, 0.0, 0.2$)\si{m} relative to the robot base.  The target location (shown as a red sphere in the figure) to be reached by the robot's end-effector tip is randomly sampled within the workspace. The task is complete when the robot finger-tip touches the target. The target speed can be one of $\{0, 0.05, 0.1, 0.2\}$\si{m/s}. For the variable speed task, the speed of the target randomly switches between these four speeds.

This task tests the stage 1 of the CLAMGen algorithm on the ability to learn a high degree-of-freedom kinematics of the robot arm via RL. It also checks if our algorithm learns a policy directly from images only (without joint and goal position). 
We compare learning from joint state only and learning from images. Observations, action and reward are specified as follows:

\begin{itemize}
    \item  \textbf{Observation 1 (Joints)}: Six joint angles $\in \mathbb{R}^6$ and the goal Cartesian coordinates $\in \mathbb{R}^3$
    \item  \textbf{Observation 2 (Image-only)}: Two RGBD images from different static cameras viewing the robot's workspace. Each image is resized to $128 \times 128$: $s_t^{i=1, 2} \in \mathbb{R}^{128 \times 128 \times 4}$.
    The goal is inferred from the image.
    \item  \textbf{Action}: change in arm joint angles i.e.  $\Delta \si{joint\_angles} \in \mathbb{R}^6 $.
    \item   \textbf{Reward}: a constant penalty of -0.01 per episode step. The maximum number of steps per episode is 100.
\end{itemize}

To leverage the benefit of fast inference time of a learned policy, we evaluate our policy on reaching a dynamic target where the goal is moving at a given static or variable speed in a random linear direction. We report the performance averaged over 200 tasks each for static speeds ($5, 10, 20$ cm/s) and a variable speed setting where the speed of the object randomly switches between the three. We measure the quality of the generated arm-trajectory by normalizing the path length with a near-optimal baseline. 
The near-optimal path is generated by interpolating from the current arm configuration to the analytical inverse kinematics (IK) of the instantaneous goal location (this approach is possible in environments with no obstacles).

Table~\ref{tab:no_obstacle_results} shows that our method is able to learn the arm reaching task either from joints or images. For either observation type, the trajectory quality is comparable or outperforms the near-optimal baseline. The inference speed of our policy is a constant that depends only on the size of the policy network (number of parameters). Even for high-dimension image inputs, the control action can be produced at $>150Hz$ since inference time of the image-based policy is $<6ms$. This enable closed-loop vision-based control.
Our method does not use any notion of motion prediction of the target and it is able to handle dynamic targets moving at a changing speed.

\begin{table}
\centering
\caption{Robot Arm Reaching to dynamic target moving at constant or varying linear velocity. We report results for a joint-only policy and an image-based policy. We measure the ratio of the path length generated by our algorithm to the path generated by a near-optimal baseline that uses analytical arm IK to move the end-effector towards the goal.}
\label{tab:no_obstacle_results}
\begin{tabular}{|c|c|c|c|c|} 
\hline
                        & Speed(cm/s) & \begin{tabular}[c]{@{}c@{}}Success\\ Rate \end{tabular} & \begin{tabular}[c]{@{}c@{}}Avg Path\\Length \end{tabular} & \begin{tabular}[c]{@{}c@{}}Avg Inference\\Speed(s) \end{tabular}  \\ 
\hline
\multirow{3}{*}{Joints} & 5     & 1.0   & 0.90175   & \multirow{4}{*}{$4.216 \times 10^{-4}$ }                          \\ 
\cline{3-4}
                        & 10    & 1.0   & 0.913     &    \\ 
\cline{3-4}
                        & 20    & 1.0   & 1.062     &   \\ 
\cline{3-4}
                        & {[}5-20]    & 1.0         & 0.9048    &            \\ 
\hline
\multirow{3}{*}{Image}  & 5           & 1.0     & 1.012 & \multirow{4}{*}{$4.906 \times 10^{-3}$ }                          \\ 
\cline{3-4}
                        & 10          & 1.0     & 1.16  &                                                                   \\ 
\cline{3-4}
                        & 20          & 1.0     & 1.20  &                                                                   \\ 
\cline{3-4}
                        & {[}5-20]    & 1.0     & 1.09  &                                                                   \\
\hline
\end{tabular}
\end{table}

\subsection{Vision-Based Arm Reaching (Obstacle-Avoidance)}
\label{sec:Fixed Obstacle}
For the obstacle avoidance task, shown in Figure \ref{fig:task_workspace} (right), we first generate two obstacles with their centers sampled in the box of $[0.35 \sim 0.65, -0.15 \sim 0.15, 0.2 \sim 0.05]$m. The obstacle can be either a cylinder or a box. If it's a cylinder, it will have diameter and height ranging from $0.06 \sim 0.09$m. Otherwise when it's a box, its width, length and height are sampled between $0.06 \sim 0.09$m separately. After obstacles are generated, we then sample a collision-free goal within the task workspace-- box of $[0.3 \sim 0.7, -0.2 \sim 0.2, 0.15 \sim 0.35]$m whose center is the obstacle sampling region. To ensure that the goal is reachable without collision, we call the IK-solver multiple times with different random seeds (start joint angles) and check if any of the solutions is collision-free. If not, we will resample the goal until a reachable one is found. In a similar fashion, the start joint angles are randomly generated to be collision-free.

The presence of obstacles adds significant complexity as the arm must learn to avoid obstacles on its path to the target location.
A task is successful if the robot is able to reach/touch the target location without colliding with obstacles in the scene.
\begin{itemize}
\item \textbf{Observation}: This includes between one and three images from different static camera views as well as the robot's joint angles and goal location. Due to possible occlusions, the robot's policy needs to learn to utilize multiple image-views to detect and avoid collision with obstacles while the keeping the goal in view.
\item  \textbf{Reward}: a constant penalty of -0.01 per episode step. After $30\%$ of total training, a collision penalty (negative reward) is introduced that is triggered when a collision occurs between the arm and an obstacle. The magnitude of this penalty grows from $0$ to $-0.1$ after $90\%$ of total training. Note that the  curriculum was chosen to start roughly at the time when training without collision penalty begins to plateau.
\item \textbf{Action}: same as the base-policy (no-obstacles) i.e. $\Delta \si{joint\_angles}$.
\end{itemize}

\begin{figure}
\centering
	
 \begin{subfigure}[h]{1\linewidth}
    \centering
    \includegraphics[width=0.975\textwidth]{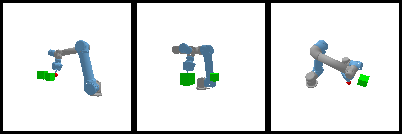}
	\caption{Three-view policy}
	\label{fig:success_three_views}
\end{subfigure}

 \begin{subfigure}[h]{1\linewidth}
    \centering
    \includegraphics[width=0.67\textwidth]{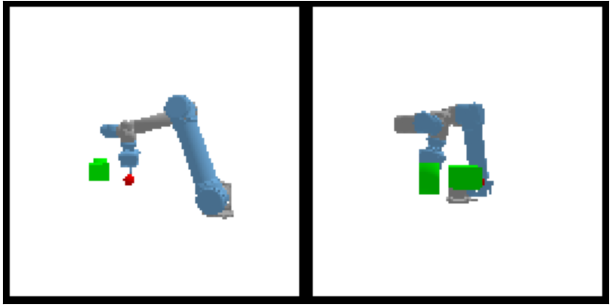}
	\caption{ Two-view policy}
	\label{fig:failure_two_views}
\end{subfigure}
\caption{\textbf{(a)}: A success case using 3 image views. The agent is able to overcome occlusion from one of the views and reach the target using the two other views \textbf{(b)}: A failure case using a two-image policy which struggles when the goal is occluded from one view. The robot aligns the end-effector with the goal in one of the views but could not achieve 3D alignment due to occlusion in the other view.}
\label{fig:result_success_failure}
\vspace{-3mm}
\end{figure}

Table~\ref{tab:obstacle_results_static_target}  shows our method is able generate collision-free arm trajectory path to reach static goals with success rate ($>0.7$) from two or more camera views. While a sampling-based approach (RRT) which has access to the full models of the objects in the scene is able to solve all the evaluation tasks, ours is able to generate arm trajectories directly from images. In addition, ours tend to generate more optimal paths compared to the RRT baseline since this is part of the reward function objective. Common failure cases include \textit{close-shave} where the robot arm slightly brushes the corner of an obstacle or inability to get to the goal due to occlusions which are still possible despite using 3 camera views.
Figure~\ref{fig:result_success_failure} shows snapshot of the terminal observation of the obstacle avoidance task. The 3-view policy is able to triangulate using the multiple views to place the end-effector at the goal location.  Figure~\ref{fig:result_success_failure}b shows a failure case for a 2-view policy which suffers because an obstacle occludes the robot from seeing the goal in that view. Although it is able to align the gripper to the goal in the unoccluded view, it still fails in that case. For similar reasons, the single view policy has a worse performance ($0.475$) due to poor observability of the environment.

\textbf{Ablation analysis} (shown in Table~\ref{tab:obstacle_results_static_target}) on the policy with two camera views reveals the relative importance of the different components of our approach.
\subsubsection{Importance of Residual learning}
As an actor-critic method, CLAMGen learns both a residual Q-function and residual policy to modulate the base Q-function and base policy action. Without residual learning, there is a $46\%$ drop in performance. We hypothesize that the greedy action of the base-policy improves exploration during learning as the agent gradually acquires the obstacle avoidance behavior. In addition the base-Q ensures that the agent's q-function always has the notion of goal-reaching.

\subsubsection{Importance of Reversible HER} Without HER, there is a $13.5\%$ drop in performance. This shows that reversible-HER complements residual learning to obtain the difficult obstacle-avoidance behavior. While HER passively improves learning by relabelling of experiences to provide more positive examples, residual learning actively improves learning by starting a much better behavior (base) policy and a good base Q-function. 
\subsubsection{Importance of View-Dropout} As demonstrated in \cite{akinola2020learning}, the view dropout mechanism is also valuable for multi-view learning. Removing this view-dropout component during training also results in a $19.5\%$ drop in evaluation performance.
\subsubsection{Importance of Collision-Dynamics Learning}
The auxiliary loss of predicting collision given a state and an action  provides more signal for policy/Q networks to learn environmental dynamics relevant to the task. Without the collision dynamics prediction, we see a $21\%$ drop in performance.
\subsubsection{Vanilla CURL with HER} This approach uses the original CURL loss\cite{srinivas2020curl} without the across-view contrastive component used by CLAMGen, does not include residual learning, and does not use collision-dynamics loss. We see a significant drop ($52\%$) in performance. This is representative of some of the approaches we tried and informative about the difficulty of obstacle avoidance task.

CLAMGen can be used for real-world applications such as closed grasping in clutter by using instance segmentation algorithms to preprocess the inputs from the camera feeds. We leave this extension as part of future work.

\begin{table}
\centering
\caption{Robot Arm Reaching to static targets with obstacle avoidance. Success rates were obtained by evaluating the best checkpoint model.
}
\label{tab:obstacle_results_static_target}
\begin{tabular}{c|c|c} 
\toprule
                      & \#Views & Success Rate \\ 
\midrule
\multirow{3}{*}{CLAMGen (Ours)} & 3 RGBD       &   0.81  \\ 
                      & 2 RGBD       &   0.76 \\ 
                      & 1 RGBD       &   0.475 \\ 
\hline
                      & \multicolumn{2}{c}{Ablation}       \\ 
\hline
No Residual             & \multirow{3}{*}{2 RGBD}       &   0.3 \\
No HER                  &                               &   0.625 \\
No View-Dropout         &                               &   0.565 \\
No Collision Learning   &                               &   0.55 \\
\hline
Vanilla CURL with HER   & 3 RGBD                   &   0.24 \\
\bottomrule
\end{tabular}
\vspace{-2mm}
\end{table}

\section{Conclusion}
\label{sec:conclusion}
This work presents CLAMGen, a closed-loop vision-based algorithm that learns a 6-DOF arm-reaching task with obstacle avoidance using multiple static camera inputs. To successfully learn this high-dimensional long-horizon task, we introduce three key ideas: residual Q-learning, 3D-awareness loss and reversible-HER. Our novel residual Q-learning approach utilizes the natural decomposition of the task to break learning into two stages: first a low-dimension arm-reaching policy and Q-function is learned using the robot's joint angles, after which the algorithm learns an image-based residual to obtain the obstacle avoidance behaviors. Our results show residual Q-learning significantly improves sample efficiency when learning from images as the base policy provides improved exploration. We also show that auxiliary losses (contrastive  loss  and  the  collision  dynamics  loss) improves performance by improving association of the camera inputs and improving the 3D understanding of the task.
We believe that our approach opens up different avenues for future work including extending CLAMGen to closed-loop grasping and applying residual Q-learning to other multi-modal robotic tasks.



\addtolength{\textheight}{-12cm}   










\end{document}